# An Explainable DistilBERT–BiLSTM–Attention Framework for Binary and Multi-Class Hate Speech Detection

**Rameesha Zia**
ORCID: 0009-0009-6717-2564
School of Computing Sciences, Pak-Austria Fachhochschule, Institute of Applied Sciences and Technology, Haripur 22620, Pakistan
rameeshaazia@gmail.com

**Muhammad Shahid Iqbal Malik***
ORCID: 0000-0001-8396-3344
School of Computing Sciences, Pak-Austria Fachhochschule, Institute of Applied Sciences and Technology, Haripur 22620, Pakistan
shahid.malik@paf-iast.edu.pk

## Abstract

Hate speech on social media poses serious risks to social harmony, mental well-being, and public safety, making its timely and accurate detection essential for content moderation systems. Most existing studies focus on binary classification, evaluated their frameworks on a single dataset, and provide limited insight into how decisions are made, which limits their real-world applicability. In addition, limited work is done on the explainability of their predictive inference. To address these challenges, this study proposes a multilevel and explainable hate speech detection framework. The proposed model integrates DistilBERT (Distilled Bidirectional Encoder Representations from Transformers) embeddings with a Bi-LSTM (Bidirectional Long Short-Term Memory) model, and an attention mechanism to capture both contextual meaning and sequential dependencies in text. To enhance trust and transparency, LIME (Local Interpretable Model-agnostic Explanations) is employed to explain model predictions by highlighting influential textual features. The framework is evaluated on two benchmark datasets using both binary and multi-class classification to examine robustness and generalization. In addition, an ablation study is presented to highlight the significance of various components of proposed framework. For binary classification, the proposed model achieves F1-scores of 96.78% on the Davidson dataset and 99.53% on the SMHS dataset. In the multi-class setting, it attains F1-scores of 97.00% and 94.99% on the Davidson and SMHS datasets, respectively, outperforming existing baseline approaches. The results demonstrate that multilevel evaluation improves the reliability that the proposed framework effectively balances performance and efficiency. This makes the framework suitable for practical hate speech moderation systems that require accurate, generalizable, and explainable decisions.

**List of Abbreviations**

| | | | |
|---|---|---|---|
| BiLSTM | Bidirectional Long Short-Term Memory | HS | Hate Speech |
| GRU | Gated Recurrent Unit | OL | Offensive Language |
| OLD | Offensive language detection | SVM | Support Vector Machine |
| SM | Social Media | RF | Random Forest |
| F-DenseCNN | Feature-Dense Convolutional Neural Network | BoW | Bag of words |
| HSD | Hate speech detection | NB | Naive Bayes |
| DL | Deep learning | LLMs | Large Language Models |
| RoBERTa | Robustly Optimized BERT Approach | SoTA | State-Of-The-Art |
| HSC | Hate speech classification | SMHS | Social Media Hate Speech |
| HSTC | Hate Speech and Target Community | FN | False Negative |
| TP | True Positive | TN | True Negative |
| FP | False Positive | NB | Naive Bayes |
| SVM | Support Vector Machines | LR | Logistic Regression |
| CNN | Convolutional Neural Network | OLD | Offensive language detection |
| BERT | Bidirectional Encoder Representations from Transformers | LR | Logistic Regression |
| XAI | Explainable Artificial Intelligence | | |

## 1. Introduction

Social media platforms provide people a freedom to share their opinions, ideas, and information with a large population [1]. This freedom plays an important role in public discussion and awareness. However, it also creates space for harmful content such as hate speech, which refers to a form of speech that targets people or groups of people based on their identity, beliefs and their background. It generates social tension and discrimination that results in real-world harm. On the other side, the internet has too much information to be moderated manually, and thus there is a need for automated systems to monitor these platforms [2].

Researchers have put forth a number of methods in recent years to identify HS using DL and ML techniques. Traditional models such as NB, SVM, and LR with manually created features were used in early researches [3]. Later studies introduced neural networks, including CNNs and LSTMs, to better capture contextual patterns in text [4]. More recently, transformer-based models such as BERT and RoBERTa have shown strong performance by learning rich contextual representations [5]. These methods have been widely tested on high-resource languages especially for English as well as low resource languages, such as Urdu [6], and have achieved promising results.

Recently, hate speech detection has attracted increasing research attention across both low- and high-resource languages. In low-resource settings, such as Urdu, several studies [7-9] have explored the detection of different forms of HS, including abusive language, violence incitement, and target identification, while highlighting challenges related to limited linguistic resources and complex language structures. In contrast, the majority of existing research focuses on high-resource languages such as English, Spanish, and French, benefiting from the availability of large annotated datasets and pretrained models. However, most of these studies primarily formulate HS detection as a binary classification task and evaluate their models on a single dataset, which limits robustness and generalization. These limitations indicate the need for more advanced HS detection systems that support multi-level classification, provide explainable predictions, and demonstrate consistent performance across multiple datasets and operational settings. To address this need, this study proposes an explainable multi-level hate speech classification framework evaluated on two English benchmark datasets.

### 1.1 Research Gap

Although significant progress has been made, several research gaps remain exist:

1. Most existing work treats hate speech detection as a binary task, which ignores different levels of severity commonly found in real social media content. Research on multi-level hate speech classification remains limited.
2. Many hate speech detection models lack explainability, making it difficult to understand why a model makes a particular decision. This reduces trust and limits their use in sensitive real-world applications.
3. Task-specific model architectures for hate speech detection are still limited. Most studies rely on standard pretrained models without adapting or optimizing them for improved contextual understanding.
4. A significant number of studies evaluate models on only one dataset, raising concerns about generalization and robustness when models are applied to unseen data.

### 1.2 Contributions

The main contributions of this study are as follows:

1. We propose an explainable multi-level hate speech detection framework that moves beyond traditional binary classification.
2. The framework integrates explainable AI using LIME to provide transparent and interpretable model predictions.
3. We introduce a hybrid architecture that combines DistilBERT embeddings, Bi-LSTM, and an attention mechanism to capture both contextual and sequential information.
4. The model is evaluated on two benchmark datasets, supporting stronger validation and better generalization.
5. Our approach achieves competitive benchmark performance, with improved accuracy and F1-score compared to standard baseline models.
6. The proposed framework provides a practical and efficient solution for automated content moderation, balancing performance with computational efficiency.

## 2. Related Work

HS, and OL are the major concerns on SM, where harmful content spreads rapidly. Researchers have proposed solutions using supervised approaches ranging from traditional ML to DL and transformer-based models. In this section, we review recent work on this topic in both English and multilingual settings, focusing on hybrid and ensemble approaches that improve detection, and handle subtle content.

### 2.1 Hate Speech Detection

Recent research on HS and OLD has shifted from simple feature-based models toward transformer-centric architectures, often strengthened by sequential models, hybrid ensembles, and strategies for low-resource languages. Across the literature, it is evident that fine-tuning a single transformer is rarely sufficient; high-performing systems usually combine either domain-adapted representations, or hybrid architectures that capture complementary lexical and contextual cues.

In an English HS context, Mamun et al. [10] investigated how rationales for sarcasm and emotive cues can improve classification performance on benchmark datasets. They enhanced the performance by identifying sarcasm-related text patterns from the HateXplain dataset [11], resulting in significant gains in F1-score compared to baseline models. It showed that subtle linguistic signals, such as sarcasm, play an important role in detecting HS in English SM. Using ensemble strategy for HSD on SM, Mubeen et al. [12] proposed a stacked model that combines SVM, RF, XGBoost, and LR to classify tweets as hate, abusive, or neutral. Their approach achieved 96% accuracy and performed better than individual classifiers. Similarly, Shilpashree and Ashoka [13] introduced an F-DenseCNN model that integrates dense CNN layers with fast word embeddings to capture both local patterns and semantic meaning. Their model outperformed baseline CNN and BiLSTM approaches on English SM datasets.

Earlier studies for English HSD showed that traditional methods such as SVM and NB were limited, mainly because they could not effectively capture the semantics of short and noisy SM text. So, the researchers shifted from traditional ML classifiers to transformers. Chapagain et al. [14] conducted a large-scale transformer study on the MetaHate dataset [15], which aggregates 36 HS datasets. By fine-tuning BERT, RoBERTa, and ELECTRA, they showed that ELECTRA achieved the best performance, with an F1-score of 89.80%, confirming the dominance of transformers over earlier RNN and CNN models on large and diverse dataset. Then, Barakat and Jaf [16] further compared LLMs with classical classifiers and found that LLMs achieve substantially better results in both binary and multi-class HS detection, mainly due to their ability to model long-range dependencies. Another study [17] suggested that even as LLMs improve accuracy, they may amplify harmful biases or generate HS themselves under adversarial conditions, a dual-use concern that requires careful mitigation strategies. Overall, research on English hate and OLD showed a strong move toward transformer-based models, with F1-scores approaching 90% on well-curated datasets, while issues of bias, interpretability, and robustness remain open challenges.

SM often contains multilingual and mixed-language content, which can be exploited to spread hate. Detecting such content in low-resource languages is difficult due to limited labeled data. To handle this challenge, the HASTIKA corpus [18] was introduced, containing 8,058 YouTube comments annotated with both binary and fine-grained hate labels. Benchmark results showed that fastText achieved 75% accuracy, while BERT reached 81%, highlighting the benefit of combining word-level and contextual representations. Recent multilingual HS research has increasingly focused on expanding beyond English by addressing data scarcity and linguistic diversity, covering languages such as Arabic, Urdu, Hindi, German, and other European languages. In Arabic SM, Mousa et al. [19] proposed a cascaded model that combines ArabicBERT with a BiLSTM layer and a radial basis function classifier. Their approach achieved strong results, with F1-scores above 89% for OLD and over 95% for HSD, demonstrating the effectiveness of integrating contextual embeddings with sequential learning models. Although with this development, the above mentioned studies did not address the issue of explainability of these black box models.

Likewise, Ahmad et al. [20] developed a human-annotated dataset for Arabic and Urdu HS and showed that transformer models, particularly XLM-RoBERTa, achieved 95% accuracy in multi-class classification, clearly outperforming existing ML and DL approaches. In leveraging DL for comprehensive multilingual HSD, Srivastava et al. [21], highlighted that models learning shared cross-lingual representations perform better than language-specific pipelines, especially in low-resource settings. Addressing data scarcity more directly, a semi-supervised GAN-based framework [22] built on mBERT and XLM-RoBERTa improved F1-scores for German, Hindi, and English using only limited labeled data, underlining the value of semi-supervised learning in multilingual scenarios.

Meanwhile, Yoo et al. [23] proposed an adaptive ensemble model that combines multiple BERT-based variants with meta-learners such as RF and SVM. Their approach consistently outperformed individual transformer on both English and Korean datasets. They achieved 85% and 89% accuracy, respectively, showing the benefit of ensemble strategies for language diversity. Similarly, Iftikhar et al. [24] used RoBERTa-large as a feature extractor with a tuned XGBoost classifier and reported strong gains on benchmark datasets, reaching 92.40% accuracy on the Davidson dataset [25] and surpassing CNN-based and standalone transformer models. Rawat et al. [26] focused on multi-class HSD and reported an F1-score of 61% with a CNN model using GloVe embeddings, supported by standard preprocessing steps such as text cleaning, concatenation, and tokenization. Summary of all the studies are presented in **Table 1.**

**Table 1:** Summary of advances in hate speech detection

| Study | Used Methods | Key findings | XAI Integration | Performance |
|---|---|---|---|---|
| [10] | BERT, Gpt-2 | Adding sarcasm cues improves detection of implicit hate | No | Acc: 74.92% |
| [12] | Stacked ensemble (SVM, RF, XGBoost, LR) | Ensemble improves hate/abusive/neutral classification | No | Acc: 96% |
| [13] | F-DenseCNN + rapid embeddings | Dense CNN + word embeddings improves hate speech prediction | No | Acc: 96.2% |
| [16] | LLM | LLMs outperform baselines | No | Acc: 88.95% |
| [14] | BERT, RoBERTa, and ELECTRA | ELECTRA outperformed all baselines | No | F1: 89.80% |
| [27] | BERT/RoBERTa | Analyzed bias from feature extraction techniques | No | F1: 68.4% |
| [18] | BERT | Hierarchical HSD | No | Acc: 80.54% |
| [19] | ArabicBERT with BiLSTM and RBF classifier | Hybrid architecture improved discrimination between offensive and hateful content | No | Recall: 92.8% |
| [20] | XLM-RoBERTa | Joint multilingual & translation approach yields high performance | No | Acc: 95% |
| [22] | mBERT + XLM-RoBERTa | Semi-supervised learning boosts F1 in low-resource languages | No | TPR: 79.03% |
| [23] | BERT-based ensemble with meta-learning (RF, SVM, WPA) | Improves precision in multilingual settings | No | Acc: ~85% (English), ~89% (Korean) |
| [24] | RoBERTa-large + XGBoost | Feature extraction + fine-tuned classifier improves accuracy | No | Acc: 97% (HSOL), 92.4% (Davidson) |
| [26] | CNN with Glove embeddings | Improves contextual understanding | No | F1: 61% |
| [7] | 1D-CNN, Urdu-BERT, Urdu-RoBERTa, BiLSTM | CNN captured unique violent patterns better than transformers | No | Acc:89.84%, Macro-F1: 89.80% |
| [8] | Urdu-Distil BERT, Urdu-RoBERTa | DistilBERT performed best for hate & target detection | No | Acc: 86.58% |
| [9] | OVB-LR, ELMo, unigram | Transparent predictions aligned with SHAP & ACME | Yes | Acc: 81.25%, Macro F1= 81.24% |
| Acc: Accuracy, F1= F1-score, BiLSTM = Bidirectional Long- and Short-term memory, OVB-LR = Optimal Variational-Bayesian Logistic Regression, ELMo = Embeddings from Language Models, HSOL= Hate speech offensive language dataset, CNN = Convolutional Neural Network, mBERT = Multilingual Bidirectional Encoder Representations from Transformers | | | | |

For Urdu-specific violence and threat content detection, Shahid et al. [7], conducted one of the earliest large-scale study on violence incitation detection in Urdu Tweets. Using a manually annotated dataset of 4,808 tweets, their proposed model (designed with a fine-tuned 1D-CNN along with unigram features) achieved 89.84% accuracy and 89.80% macro-F1, outperforming several transformer and traditional ML baselines. Extending Urdu HS research, Malik et al. [8] introduced the first large-scale Nastaliq Urdu HS and target classification corpus based on Facebook data. By fine-tuning Urdu-RoBERTa and Urdu-DistilBERT, they showed that DistilBERT achieved strong results, with 86.58% accuracy for binary HSD and competitive performance for multi-class target identification, demonstrating that lightweight transformers are effective even for complex scripts. Finally, Nazarova et al. [9] introduced an interpretable framework for English threat detection that combines probability estimation with feature-weight modeling. Their approach outperformed several traditional classifiers and produced explanations using SHAP and ACME XAI, showing that effective performance and interpretability can be achieved together.

Taken together, these studies point to a clear shift from traditional feature-based classifiers to transformer-based and hybrid models. While most research emphasized improving detection accuracy, only limited attention has been given to combining deep contextual understanding with transparent and interpretable explanations, leaving an important gap for future work.

**3. Dataset Description**

In this study, we used two publicly available datasets that are widely used in HS and OL detection tasks. Both datasets were used in two ways. First, we trained our proposed framework in a binary setting where the task was to decide whether a post is hateful or not. Second, we used their original label structures for multi-class experiments, where the goal was to differentiate between different types of hate and not-hate content. This allowed us to test how well the model behave in both simple and more detailed classification settings. Details of these datasets are discussed in following subsections.

**3.1 Davidson dataset**

The first dataset we have used in this study is the Davidson dataset [25], which contains about 24,800 English tweets collected from Twitter using hate-related keywords. We used this dataset in two forms. For multi-class classification, we kept the original three labels that are ‘hate speech’, ‘offensive language’, and ‘neither’. For binary classification, we merged the HS and OL labels into “hate” class, while the “neither” class was treated as “not hate”. This setup allows the same data to support both fine-grained and binary classification tasks. Word cloud of nouns for the Davidson dataset is shown in ***Fig. 1***.

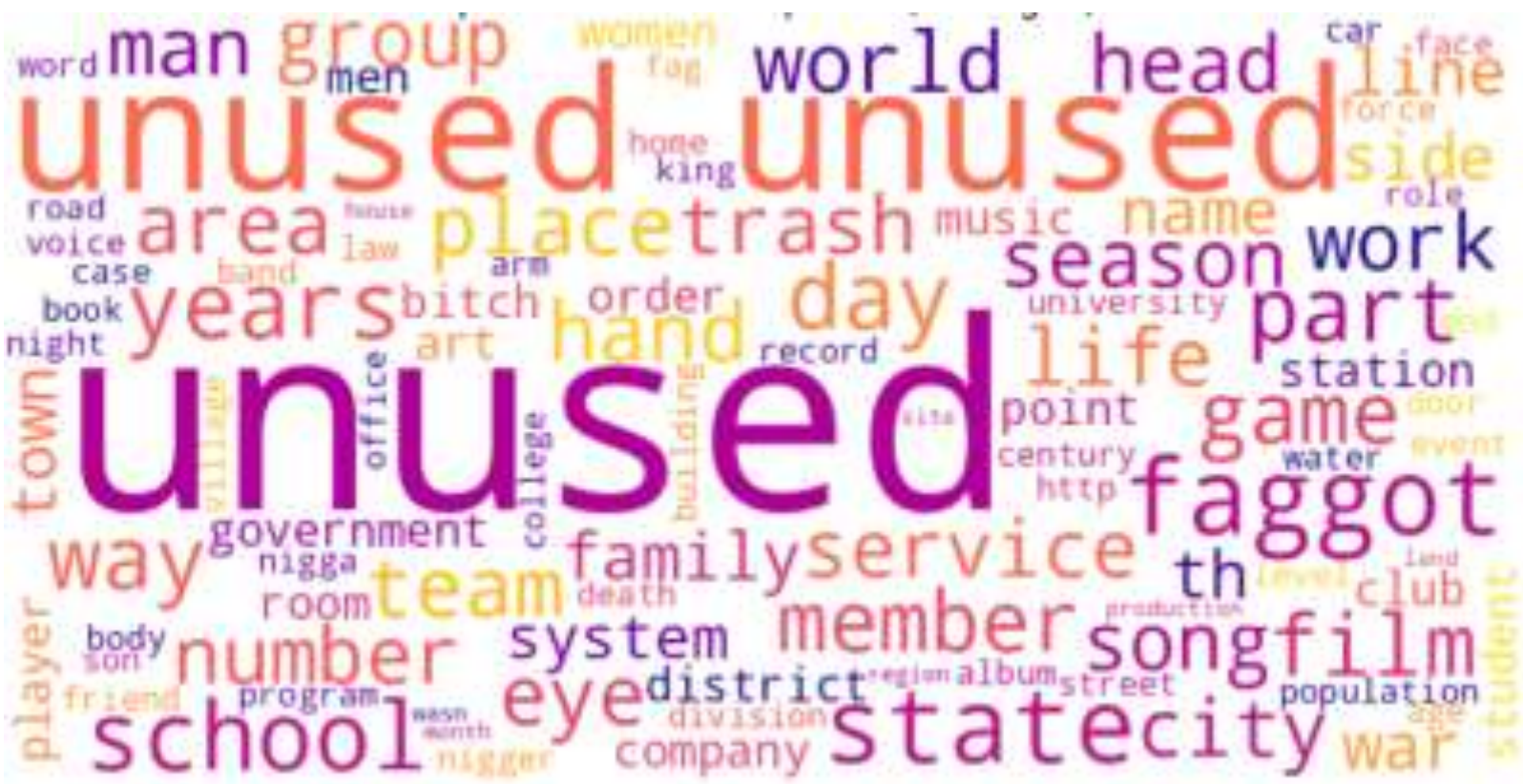

**Fig. 1**: Word cloud of nouns in the Davidson dataset

**3.2 SMHS dataset**

The SMHS dataset [28] is the second one, that we have used in this study. It has a total of 13,000 samples divided into six classes for HSD. The classes of this dataset are: offensive, anti-state, anti-religion, sexist, racist, and others, each having enough instances to aid in better model training and learning processes. We used this dataset for both multi-class and binary classification tasks. In the multi-class setting, all original labels (six) were kept. For the binary setting, all classes related to hate, abuse, or harmful behavior were grouped into a single "hate" category, while the remaining content was labeled as "not hate". This lets us test how well the model handle both detailed category prediction and binary hate speech detection using the same dataset. Word cloud of SMHS dataset is shown in ***Fig. 2*.**

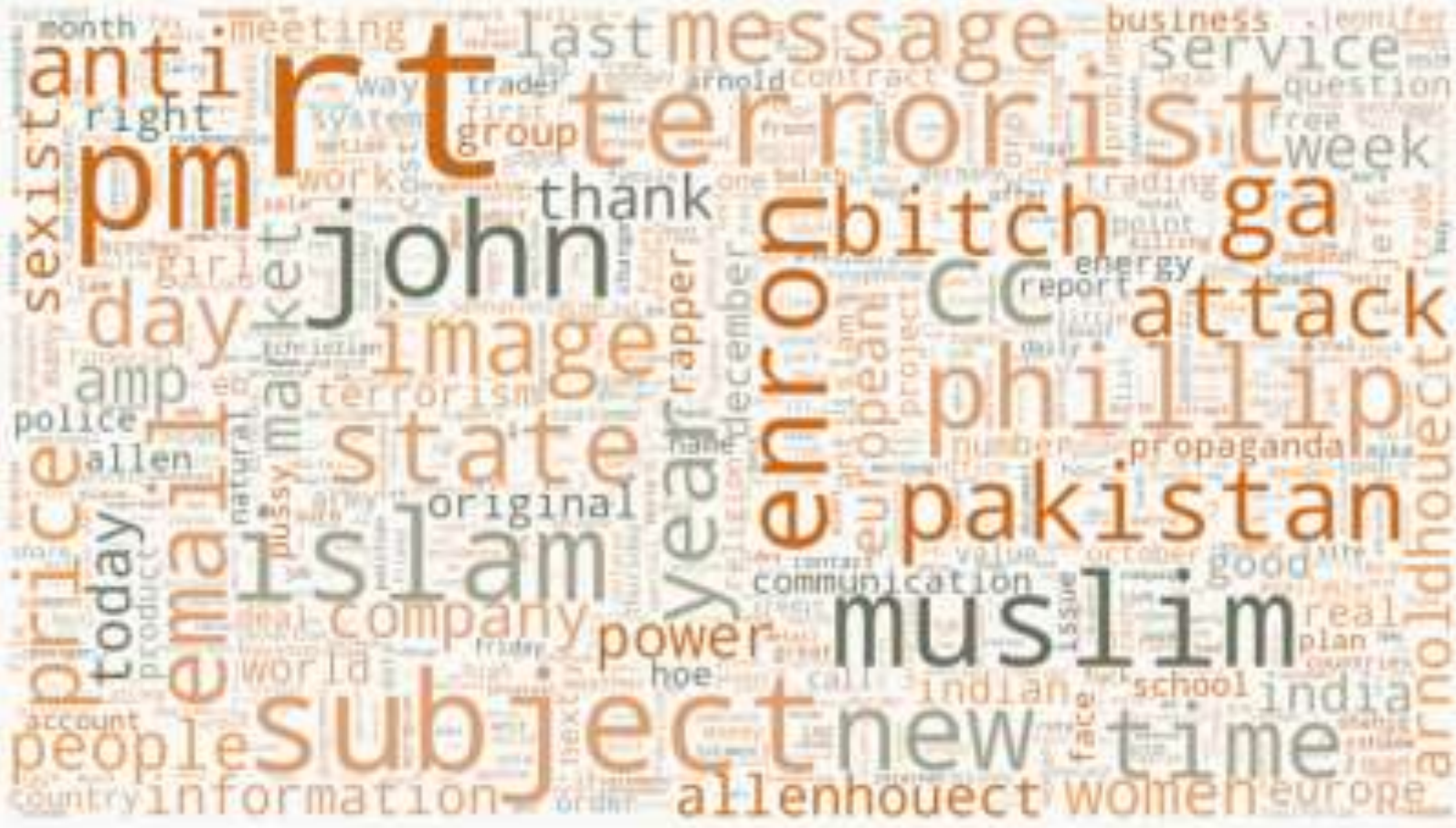

**Fig. 2**. Word cloud of nouns in the SMHS dataset

## 4. Methodology

This section explains the overall workflow followed in this study, from data preparation to model development and evaluation. The aim is to clearly describe how the raw SM text was transformed into a suitable form for learning, and how the proposed hybrid model was applied for both binary and multi-class HSC. An overview of the complete workflow of the proposed system is illustrated in **Fig. 3,** which shows the main components and their interactions.

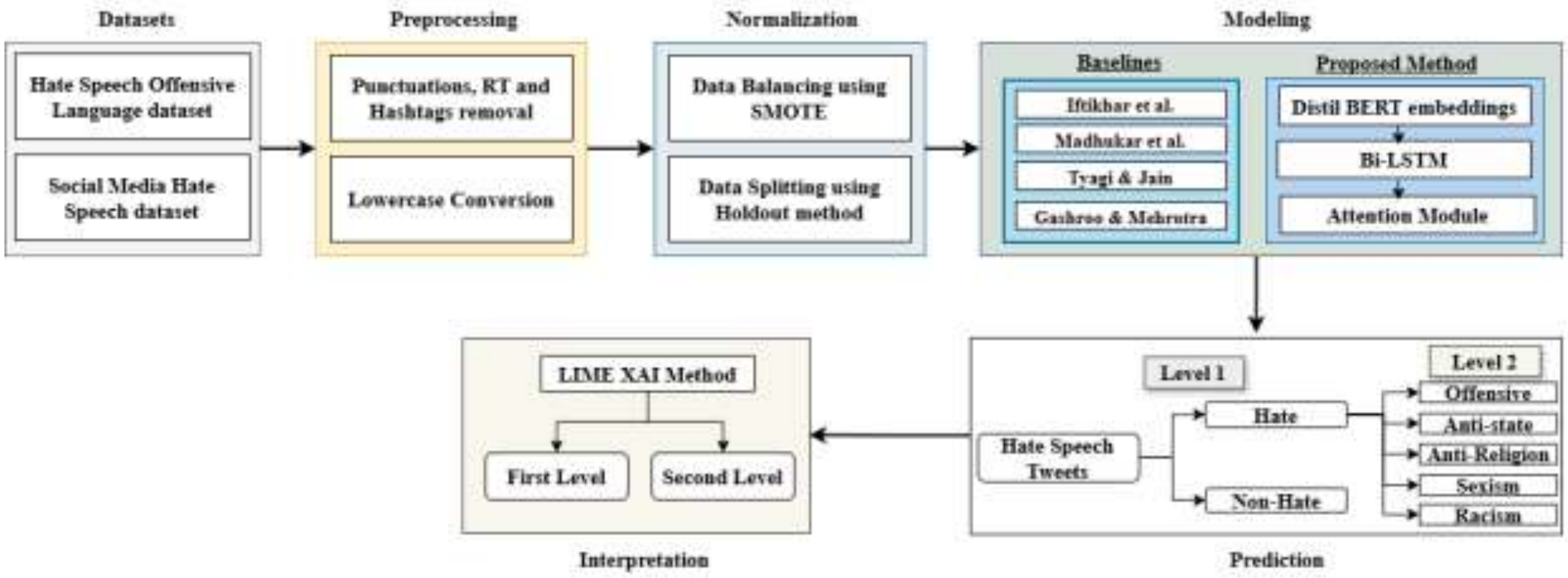

**Fig. 3**: Working procedure of the proposed method

### 4.1 Preprocessing

SM text is often noisy and unstructured, which can negatively affect model performance if not handled carefully. To address this, a series of text cleaning steps were applied to normalize the data and remove irrelevant elements while preserving the core meaning of the content:

1. All text was first converted to lowercase to keep the representation uniform. Hashtags were eliminated to reduce noise in the text.
2. Usernames, retweet markers, and excessive exclamation symbols were also removed because they do not add useful information for identifying hate content.

3. Numbers and punctuation marks were removed as they rarely contribute to the meaning of hateful content.
4. After cleaning, extra spaces created during the process were fixed to maintain proper word separation.

These steps helped simplify the text while keeping its original meaning, allowing the model to focus on meaningful linguistic patterns rather than irrelevant symbols or formatting.

### 4.2 Distill-BERT Embeddings

After preprocessing the dataset, the textual data is transformed into dense vector representations (known as embeddings), and a pre-trained transformer model is used for this purpose. The embeddings are the numerical representations of text that capture the contextual relationships and semantic meaning of words. The first layer of the hybrid multilevel DL model is a DistilBERT in which raw text is tokenized into sub-word units that the DistilBERT model can understand. This embedding model [29] is chosen for its efficiency in terms of speed and resource utilization, while maintaining a high level of performance close to BERT. It is a smaller, faster, and lighter version of the BERT model, pre-trained on large text corpora, and can generate high-quality contextual embeddings for various NLP tasks. Each token is represented by a dense vector of 768 dimensions. This high-dimensional representation captures intricate semantic and syntactic information about the token in the context of the entire sequence. The tokenized tweets, including special tokens [CLS] and [SEP], were passed as input to the DistilBERT model. Each token in the sequence is mapped to a unique integer index based on the tokenizer's vocabulary. DistilBERT consists of multiple layers of transformers, where each layer applies self-attention and feedforward neural networks to the input tokens. This process generates contextual embeddings for each token by considering its relationship with other tokens in the sequence. The output corresponding to the [CLS] token is extracted as the aggregate representation of the entire tweet.

### 4.3 Bidirectional-LSTM

The output embeddings from DistilBERT are passed to a Bi-LSTM layer as shown in **Fig. 4.** The role of the BiLSTM is to further enhance the sequential nature of the text as it helps to understand the sequence of data from both directions i.e., forward and backward. To maintain a high level of detail, the layer is configured with a hidden dimension of 256, producing a rich output vector for every step in the sequence. This enables the model to capture long-range dependencies that may not be fully preserved by transformer embeddings alone. By combining the contextual representations generated by DistilBERT with the sequential learning capability of the Bi-LSTM, the model benefits from both global context and ordered word information. This hybrid design improves the model's ability to identify subtle linguistic patterns related to HS, which often rely on sentence structure rather than isolated keywords. The output of the Bi-LSTM layer is then forwarded to the attention mechanism for further refinement before classification.

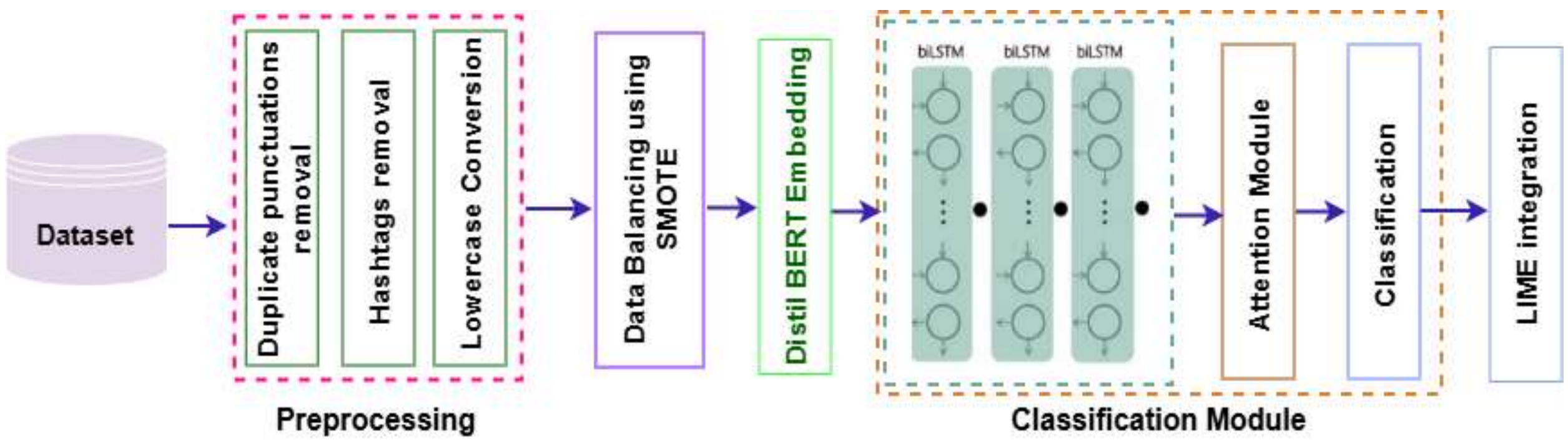

**Fig. 4**: Description of the Proposed Architecture

### 4.4 Attention Module

Next, the output of the Bi-LSTM layer is passed through an attention module. The attention layer is designed to assign varying levels of importance to different tokens in the sequence, allowing the model to focus on the most relevant parts of the input. This is particularly useful in tasks like text classification, where some parts of the sequence (e.g., certain words or phrases) might carry more significant information than others. The attention module in the proposed model uses a linear layer that computes a scalar attention score for each time step in the Bi-LSTM output. The attention scores are then normalized using the SoftMax function, transforming them into a probability distribution that sums to 1. These scores are used to compute a weighted sum of the Bi-LSTM outputs, creating a context vector. This context vector represents a summary of the entire input sequence, where tokens with higher attention weights contribute more to the final representation.

### 4.5 Classification

Finally, the context vector generated by the attention module is passed through the classifier layer to make the final predictions for both binary and multiclassification. This is a fully connected (linear) layer that takes the context vector, which has a hidden dimension of 2 due to the Bi-LSTM, and outputs a set of logits, representing the predicted scores for each of the number of label classes. The classification layer transforms the context vector into the final class probabilities.

### 4.6 LIME Integration

After training and validation of the proposed model, we used LIME method to explain and understand how the model was making its predictions. LIME is an XAI technique [30] that can be utilized to explain each of the predictions made by the proposed model. No matter how complex the model is, LIME can explain the reasoning behind each individual prediction. By combining interpretability with predictive performance, LIME not only facilitates error analysis and debugging but also supports ethical AI practices. In the domain of HSD, this is particularly critical, as it ensures that the model's decisions are not only accurate but also explainable and fair. This capability to balance complexity with clarity makes LIME a powerful tool for enhancing the usability and accountability of advanced AI systems. For instance, if a user posts a comment that the model classifies as offensive, LIME can highlight the key phrases or words that led to this classification. This not only helps the model in understanding the behavior but also provides valuable feedback for users who may want to know why their content was flagged.

### 4.7 Hyperparameters Tuning

The proposed model was fine-tuned with appropriate hyperparameters to achieve the best results and the detail of hyperparameters are shown in **Table 2**. The model was trained and tested for 10, 20, and 30 epochs to analyze their impact on learning and performance. Different batch sizes of 16, 32, and 64 were tested, with the best results achieved using a batch size of 16. Input text was tokenized using the DistilBERT tokenizer with maximum sequence lengths of 32, 64, and 128, ensuring the model processed the text efficiently while retaining key information. We split the dataset into three sets; 70% training, 15% testing and 15% validation sets. Among different optimizers and learning rates, the AdamW optimizer with a learning rate of $2 \times 10^{-5}$ gave the best performance, helping the model update weights smoothly. The selected hyperparameters allow the model to generalize well on both datasets. The highest performance was observed when 30 epochs are applied, with loss, accuracy, precision, recall, and F1-score monitored for both training and validation sets. The cross-entropy loss function was used to measure model performance.

**Table 2**: Details of hyperparameters for the Proposed Model

| **Parameters** | **Values** |
|---|---|
| Optimizer used | RMSProp, Adam, AdamW |
| Learning Rate | 2 × 10-5, 2×10−2, 2×10−3 |
| No. of Epochs | 10, 20, 30 |
| Batch Size | 16, 32, 64 |
| Max Sequence Length | 32, 64, 128 |

## 5. Experimental Setup

The initial setup was a laptop with Intel core-i5 ($8^{th}$ generation) and 8-GB RAM specifications which provided a good baseline to start the initial development and testing. Realizing the limitations of this local hardware and aiming at increasing computational performance, later experiments were run on Kaggle with a T4 GPU on its cloud. A switch to this high-performance infrastructure allowed this model to be fine-tuned; the training process was accelerated by the GPU, which had a significant decreasing effect on iteration time and provides a more robust platform to continue the experiments. This setup also enabled faster experimentation, reducing training times while maintaining model performance.

### 5.1 Baseline Models

This section discusses the baseline models chosen for comparison with the proposed approach. To ensure a fair comparison, we selected three benchmark studies for the Davidson dataset and one study for the SMHS dataset. These baselines represent commonly used hybrid and DL models for hate and OLD tasks.

#### 5.1.1 Baselines for Davidson dataset

For the Davidson dataset, three baseline approaches are chosen:

1. The first baseline is the study [24], in which the authors integrated two state-of-the-art techniques, i.e. RoBERTa and XGBoost models and demonstrated significant performance.
2. The second baseline is the study [31], where a CNN followed by an LSTM model is used for hate and OLD tasks. In this model, the CNN captured local textual patterns, while the LSTM models the sequential dependencies in the text.
3. The study [32] is used as the third baseline, in which the authors proposed a hybrid LSTM-CNN based approach. The sequential information is learned using LSTM, and then the CNN extracts discriminative local features.

#### 5.1.2 Baseline for SMHS dataset

For the SMHS dataset, only one baseline is available:

1. This study [33] utilized DistilBERT embedding model and further fine-tuned it to achieve the optimum performance.

### 5.2 Evaluation Metrics

Standard classification metrics are utilized here to evaluate the performance of proposed and baseline approaches for the said task. These metrics are commonly used in the related HS and OLD tasks. The metrics are: Accuracy, Precision, Recall, and F1-score. Brief description and formulas of the metrics is presented in **Table 3**.

**Table 3**: Evaluation Metrics

| Metrics | Description | Formula |
|---|---|---|
| **Accuracy** | Overall correctness of predictions | $\frac{TP + TN}{TP + TN + FP + FN}$ |
| **Precision** | Correctly predicted hate samples among predicted hate | $\frac{TP}{TP + FP}$ |
| **Recall** | Correctly detected hate samples among actual hate | $\frac{TP}{TP + FN}$ |
| **F1-score** | Balanced measure of precision and recall | $2 \times \frac{Precision \times Recall}{Precision + Recall}$ |
| **True Positive (TP):** The model correctly predicts a positive class, **TN (True Negative):** The model correctly predicts a negative class, **FP (False Positive)**: The model predicts positive, but the actual class is negative, **FN** (**False Negative):** The model predicts negative, but the actual class is positive | | |

## 6. Results and Analysis

This section presents the experimental results of the proposed model on the Davidson and SMHS datasets and its comparison with standard baseline models. Performance is evaluated for both binary and multiclass classification tasks, followed by an ablation study to assess the contribution of individual model components. Furthermore, LIME is employed to interpret model predictions and analyze the influence of key words on classification outcomes.

### 6.1 Ablation Study

An ablation study was performed to examine the impact of different components of the proposed model. The study focused on the impact of the number of Bi-LSTM layers, the use of attention module, residual concatenation, and SMOTE. Each setting was evaluated using the F1-score, number of trainable parameters, and training time per batch. The results for all configurations are reported in **Table 4** for both the Davidson and SMHS datasets.

In the first setting (study 1 & 4), we used DistilBERT embeddings followed by a single Bi-LSTM layer, without (attention, residual concatenation, and SMOTE) modules. This configuration achieves an F1-score of 93.65% on the Davidson dataset and 98.12% on the SMHS dataset. The number of trainable parameters remains 67.48 million, and the training time per batch is 20 seconds for both datasets. In the second setting (study 2 & 5), residual concatenation is added while keeping the rest of the model unchanged. This leads to a slight improvement in the F1-score, reaching 93.74% on the Davidson dataset and 98.44% on the SMHS dataset. The number of trainable parameters remains the same, as residual connections do not introduce new learnable weights. The training time shows a small increase due to the additional computation during feature combination.

The third setting (study 3 & 6) increases the number of Bi-LSTM layers to two and includes both attention and residual concatenation. In this case, the F1-score slightly decreases on the Davidson dataset to 92.92%, while it improves on the SMHS dataset to 98.56%. The number of trainable parameters increases to 67.87 million, reflecting the added Bi-LSTM layer. The training time remains close to earlier settings, showing that the increase in model depth has a limited impact on computational cost. Across all these settings, the number of trainable parameters remains nearly the same, as DistilBERT contributes the majority of the parameters. The final setting corresponds to the full proposed model. It uses SMOTE, a single Bi-LSTM layer, an attention module, and residual concatenation. With this setup, the model gave the best performance among others on both datasets. It reaches an F1-score of 96.78% on the Davidson dataset and 99.9% on the SMHS dataset. The number of trainable parameters does not change much compared to earlier settings, and the training time per batch also remains the same. The training time across all experiments stays low and stable. This is mainly because DistilBERT is used as the embedding layer. DistilBERT is smaller and faster than many other transformer models. Most of the parameters come from this base model, while the added attention and residual components do not add much extra computation. As a result, the model stays efficient during training.

**Table 4:** Ablation study of the proposed method

| **Ablation Study No.** | **SMOTE (YES/NO)** | **No of Bi-LSTM** | **Attention Module (YES/NO)** | **Residual Concatenation (YES/NO)** | **Dataset (F1-Score (%))** | **Trainable Parameters (Millions)** | **Training Time Per Batch (Seconds)** |
|---|---|---|---|---|---|---|---|
| 1 | NO | 1 | NO | NO | Davidson dataset (93.65) | 67.48 | 19.99 |
| 2 | NO | 1 | NO | YES | Davidson dataset (93.74) | 67.48 | 20.19 |
| 3 | NO | 2 | YES | YES | Davidson dataset (92.92) | 67.87 | 20.19 |
| 4 | NO | 1 | NO | NO | SMHS dataset (98.12) | 67.48 | 20.19 |
| 5 | NO | 1 | NO | YES | SMHS dataset (98.44) | 67.48 | 20.24 |
| 6 | NO | 2 | YES | YES | SMHS dataset (98.56) | 67.87 | 20.24 |
| Proposed Model | YES | 1 | YES | YES | Davidson dataset (96.78) | 67.48 | 20.17 |
| Proposed Model | YES | 1 | YES | YES | SMHS dataset (99.9) | 67.28 | 19.55 |

In summary, the ablation results show that adding attention, residual concatenation, and SMOTE helps improve performance. These improvements are achieved without increasing the model size or training time in a noticeable way. This confirmed the proposed model is both effective and computationally efficient.

### 6.2 Performance Analysis for Binary Classification

This section describes the experiments performed to test the performance of proposed model and its comparison with standard baseline models for binary classification task. The training and validation performance of the proposed model is analyzed using loss and F1-score curves on both datasets. These curves help in understanding how well the model learns during training and how it performs on unseen data. **Fig. 5** shows training and validation losses and F1-scores for both datasets for 15 epochs. For the Davidson dataset, the training loss shows a clear downward trend across the epochs as shown in **Fig. 5** (a). In the early epochs, the loss is relatively high, which means the model is learning the basic patterns in the data. As training progressed, the loss gradually decreases and becomes very small reaching 0.01%. The validation loss follows a similar pattern and remains close to the training loss. A similar trend is observed for the SMHS dataset as shown in **Fig. 5** (c). The training loss decreases sharply in the first few epochs and remains very close to zero afterward. The validation loss stays low throughout validation, reaching zero at 13$^{th}$ epoch. This behavior indicates that the model is learning in a stable manner and is not overfitting. When both training and validation losses are low, it means the prediction error is small and the model is making correct decisions for most samples.

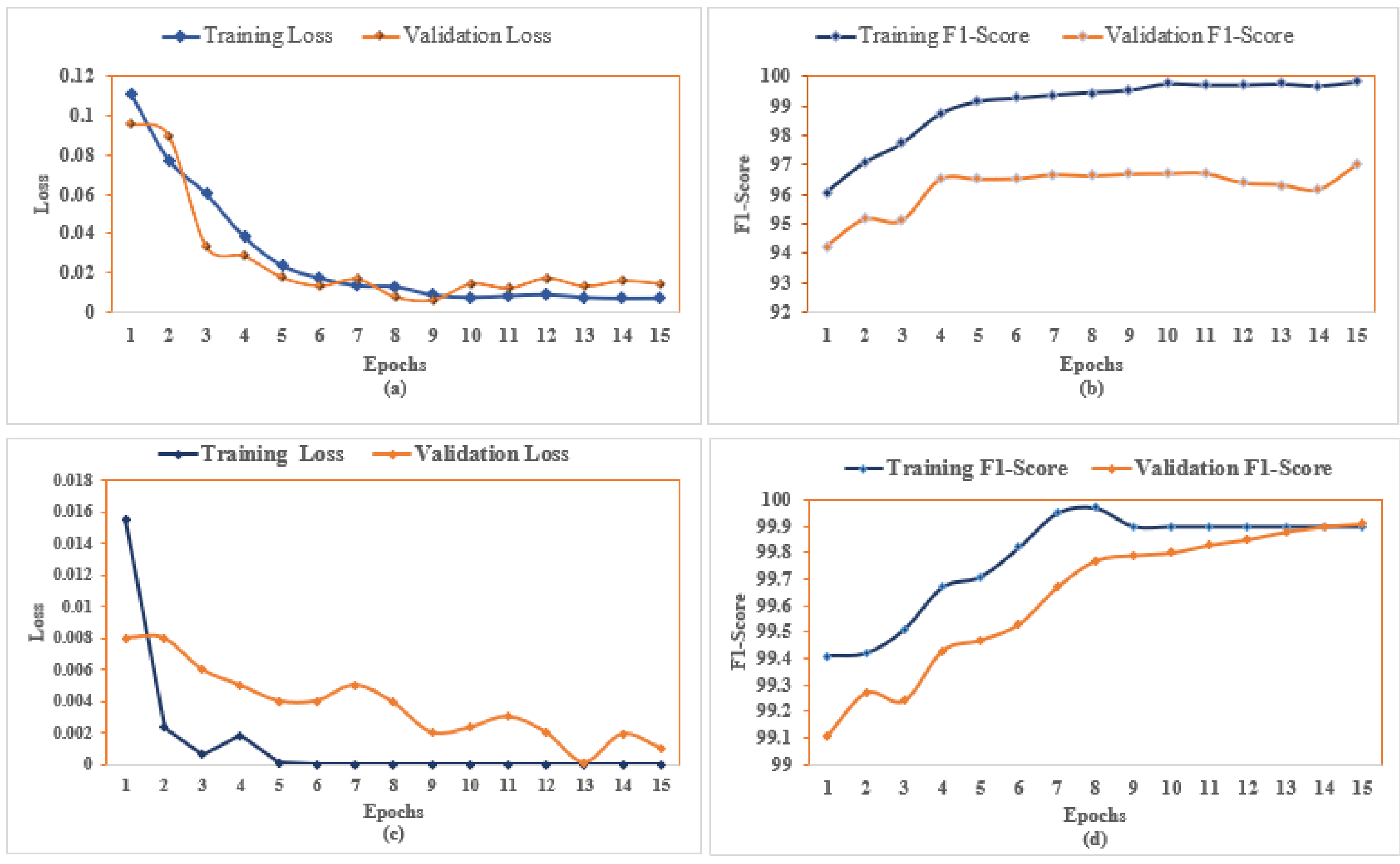

**Fig. 5:** Training and validation performances for binary classification (a) Loss on Davidson dataset (b) F1-Score on Davidson dataset (c) Loss on SMHS dataset (d) F1-Score on SMHS dataset

The F1-score curves for the Davidson dataset further support this observation. As shown in **Fig.** ***5*** (b), the training F1-score increases steadily from 96% and reaches nearly 99% by the 15th epoch. The validation F1-score also improves with increasing number of epochs and approached at 96.78%. The small difference between training and validation F1-scores shows that the model performs consistently on both training and validation data. This confirms that the learned features generalize well to unseen samples. The curves for F1-score for the SMHS dataset is shown in **Fig. 5** (d). We can see that the training F1-score increases quickly and reaches almost 100%, while the validation F1-score gradually rises and stabilizes at 99%. These high values indicate that the model correctly identifies both hate and non-hate classes with very high performance. The close alignment of the training and validation F1-score curves suggest that the model does not memorize the training data and maintains strong generalization capability. Overall, low loss values indicate the difference between predicted labels and true labels is minimal, while high F1-scores reflect a good balance between precision and recall. The consistent improvement and stability of both loss and F1-score across epochs demonstrate that the proposed model is well-trained and achived benchmark performance on validation.

The confusion matrix helps in understanding how the model makes decisions for each class. It shows how many samples are correctly classified and where the model makes mistakes. For the Davidson dataset, the model correctly predicted 3344 samples as hate (that is the TP) as shown in **Fig. 6 (a),** which shows that the model has learned strong patterns related to HS. Whereas, the model correctly classified 3398 samples as not-hate, which represents the TN. This high TN value indicates that the model is also effective at recognizing not-hate tweets. At the same time, there are 97 FP (means not-hate content is incorrectly labeled as hate). This usually happens when strong or aggressive words appear in a neutral context. In addition, 150 hate samples are misclassified as not-hate (i.e. FN). These cases often occur when HS is indirect, sarcastic, or expressed using subtle language, making it harder for the model to detect.

The results are more consistent for the SMHS dataset, as shown in **Fig. 6 (b).** The model correctly predicted 3365 samples as not-hate (TN) and 3362 samples as hate (TP). These very high values show that the model separates the two classes clearly. Only 1 FP is observed, meaning the model approximately do not mislabel normal content as hate. In addition, there are only 4 FN, which indicates that very few hate samples are missed. This strong performance exhibits the learning of clearer class boundaries and better annotation quality in the SMHS dataset.

Overall, the high values of TP and TN scores for both datasets show the effectiveness of the proposed model in identifying hate and not-hate speech content. The low FP scores show the model's ability to avoid over-predicting HS, and relatively low FN scores show that most of the HS is indeed identified. These points confirm the high F1-scores and show that the proposed model provides balanced and efficient classification performance.

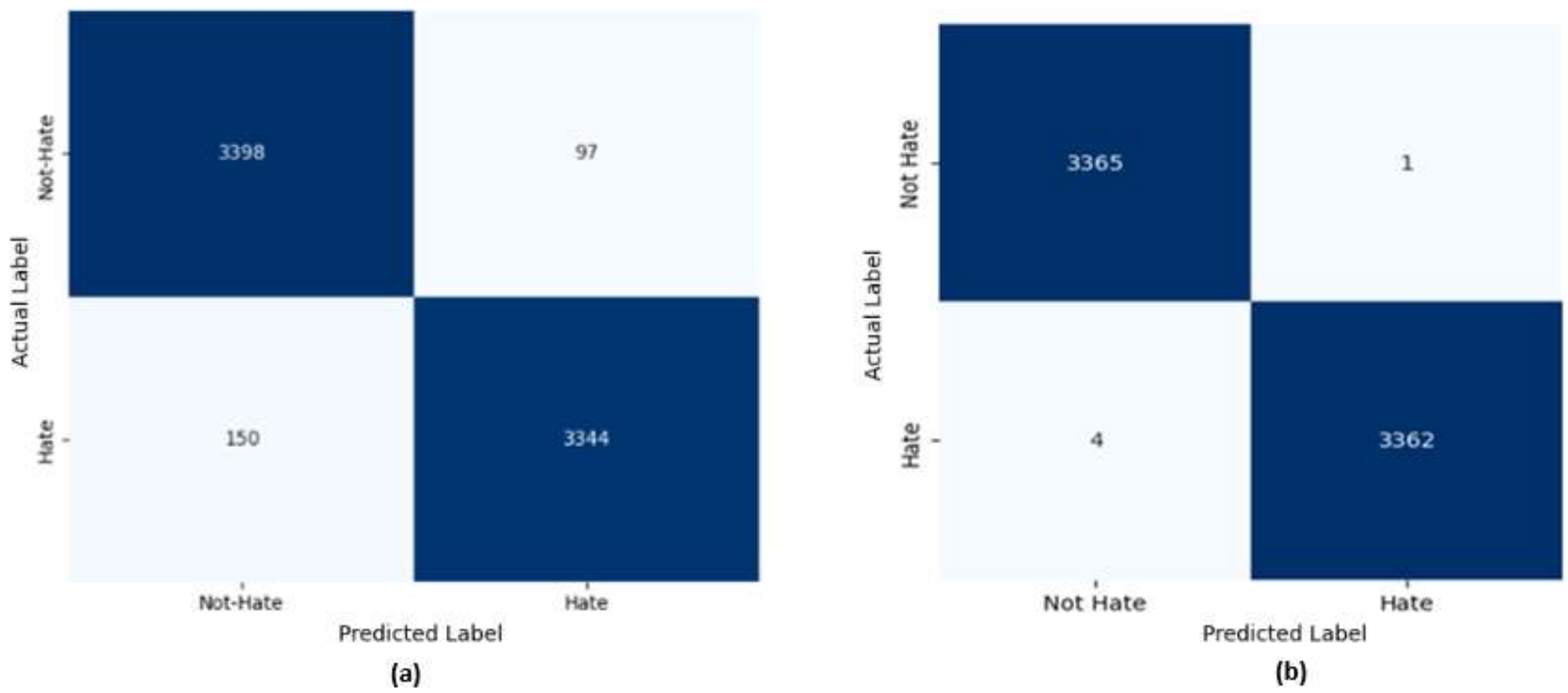

**Fig. 6:** Confusion matrix on (a) Davidson dataset, (b) SMHS dataset

We also compared our work with earlier studies (baseline models) that used the same datasets to evaluate binary classification performance. The comparison results are summarized in **Table 5**. On the Davidson dataset, previous studies [24, 31] reported accuracy values between 90.05% and 94.50%, with F1-scores ranging from 92.09% to 93%. These models utilized hybrid DL architectures but exhibited limitations in balancing precision and recall. In contrast, the proposed model achieved an accuracy of 96.78% and an F1-score of 96.44%, showing a clear improvement over existing baseline approaches. The higher precision and recall values indicate that the model is better at identifying HS while reducing misclassification.

**Table 5:** Comparison of classifiers for binary classification

| Dataset | Models | Accuracy (%) | Precision (%) | Recall (%) | F1-Score (%) |
|---|---|---|---|---|---|
| **Davidson dataset** | **[24]** | 94.50 | 92.89 | 93.00 | 92.78 |
| | **[31]** | 90.05 | 91.78 | 88.90 | 92.09 |
| | **[32]** | 93.99 | 92.76 | 91.84 | 93.00 |
| | **Proposed** | **96.78** | **97.18** | **95.71** | **96.44** |
| **SMHS dataset** | **[33]** | 90.03 | 92.66 | 92.65 | 92.76 |
| | **Proposed** | **99.99** | **99.90** | **99.76** | **99.78** |

For the SMHS dataset, the baseline model achieved an F1-score of 92.76%, which reflects good performance. However, the proposed model significantly improves the baseline results, achieving 99.99% accuracy and an F1-score of 99.78%. This large performance gap shows that the proposed model is more effective at learning clear decision boundaries between hate and not-hate classes. Overall, the results demonstrate that the proposed model consistently outperforms baseline methods on both datasets by achieving higher accuracy and better balance between precision and recall measures.

### 6.3 Performance Analysis for Multiclass Classification

After completing the binary classification experiments, we further evaluated the proposed model on multi-class classification for both the Davidson and SMHS datasets. This analysis helps in understanding how well the model learns complex class boundaries when more than two classes are involved.

Training and validation loss tell us how well the model learns during training and how it performs on unseen data. The loss curves for Davidson and SMHS datasets are shown in **Fig. 7 (a) and (c)** respectively. For both datasets, the training loss decreases sharply in the initial epochs and then gradually stabilizes at zero, indicating that the model learns meaningful patterns early during training. The validation loss also follows a decreasing trend and remains close to the training loss, which suggests stable learning and limited overfitting. As shown in **Fig. 7**(a), the validation loss for the Davidson dataset reaches 0.04, while for the SMHS dataset (in **Fig. 7** (c)), the validation loss settles around 0.13. Although the validation loss remains higher than the training loss for SMHS dataset, the overall downward trend indicates consistent learning. The gap between training and validation loss is acceptable and does not increase significantly, suggesting that the model maintains generalization despite the higher complexity of the SMHS dataset.

The training and validation F1-scores explain how well the model balances precision and recall during learning. The F1-score curves for the Davidson and SMHS datasets are shown in **Fig. 7** (b) and **Fig. 7** (d), respectively. In early epochs, the training F1-score increases rapidly and stabilizes at a high value close to 99%, showing strong learning performance for Davidson dataset. The validation F1-score also improves steadily and closely follows the training curve reaching 97%, indicating balanced precision and recall across multiple classes. Similarly, for the SMHS dataset, the training F1-score shows a continuous improvement and approaches 99% as training progresses. The validation F1-score remains stable at 94.99% with small fluctuations across some last epochs, which is expected in multi-class classification tasks with complex data. Despite these minor variations, the validation F1-score remains high, confirming that the proposed model generalizes well and maintains consistent performance across different classes. Overall, both the loss and F1-score curves confirm that the proposed model learns effectively, avoids overfitting, and performs consistently well on both datasets in the multi-class classification setting.

We also showed confusion matrices for both datasets as it provides a clear picture of how the model predicts each class and where small mistakes occur. It helps us understand not only the overall performance but also the behavior of the model for individual classes. The confusion matrices for Davidson and SMHS datasets are shown in **Fig. 8(a) and (b)** respectively**.**

For the Davidson dataset as shown in **Fig. 8(a)**, when we consider one class at a time, the number of TP is high. For example, for the Hate class, 1972 samples are correctly predicted as hate (TP), whereas 4675 is the TN value for the Hate class which is quite large. Likewise, FP occur when "offensive and neither" tweets are wrongly predicted as hate, such as 69 "offensive and 979 neither" tweets. The FN are actually hate tweets but predicted as "offensive and neither" tweets, such as 117 predicted as offensive and 783 as neither. Similar patterns are seen for the Offensive and Neither classes, where TP values remain high and FP and FN values are quite low. This shows that the model is able to correctly capture class-specific patterns, with errors mainly caused by overlap in language between hate, offensive, and neutral content.

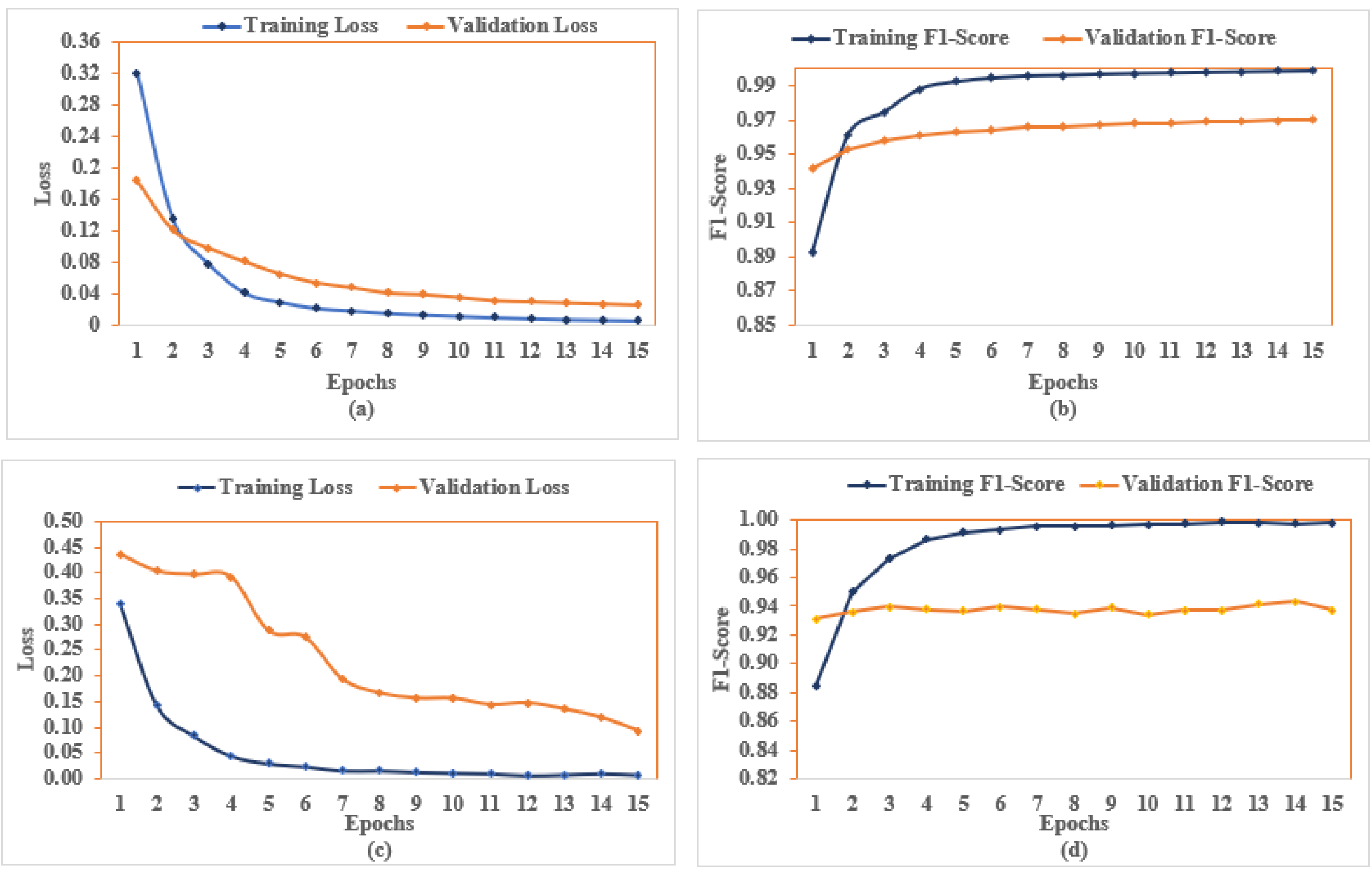

**Fig. 7:** Training and validation performances for multi-class classification (a) Loss on Davidson dataset (b) F1-Score on Davidson dataset (c) Loss on SMHS dataset (d) F1-Score on SMHS dataset

For the SMHS dataset, the TP values are very high for all six classes as shown in **Fig. 8 (b)**. For example, the model correctly predicts 813 anti-religion, 223 anti-state, 812 normal, 765 offensive, 737 racism, and 724 sexism samples as TP. The FP and FN are very small in comparison. For instance, only a few racism samples are misclassified as offensive or sexism classes, and a small number of sexism samples are predicted as racism. The TN remain high because samples from other classes are mostly classified correctly. Overall, the high TP and TN values and the low FP and FN values indicate strong classification performance. Most of the misclassifications occur among the closely related categories, which explains why FP and FN are not zero. This behavior is expected in multi-class hate and OLD tasks. In addition, overall it confirms that the model is learning meaningful and reliable patterns.

We also compared proposed model with baseline studies that used the same datasets for multiclassification task, and the results are presented in **Table 6**. The results indicate a substantial improvement in the proposed model's performance metrics, particularly in F1-score, which increased to 97% and 94.99% for Davidson and SMHS datasets respectively, surpassing the benchmark studies.

For the Davidson dataset, baseline studies achieved accuracies between 88.49% and 93.10% and F1-scores between 90.05% to 93.60%. These studies utilized combinations of CNN, LSTM, or RoBERTa with ensemble classifiers, demonstrated limited performance either by insufficient sequential modeling or by the lack of an attention mechanism, which restricted their ability to focus on the most discriminative parts of the text. In contrast, our model uses

DistilBERT embeddings with BiLSTM and attention module, captures both contextual and sequential information, resulting in a higher accuracy of 96.09%, balanced precision (96.78%) and recall (96.07%), and an F1-score of 97%.

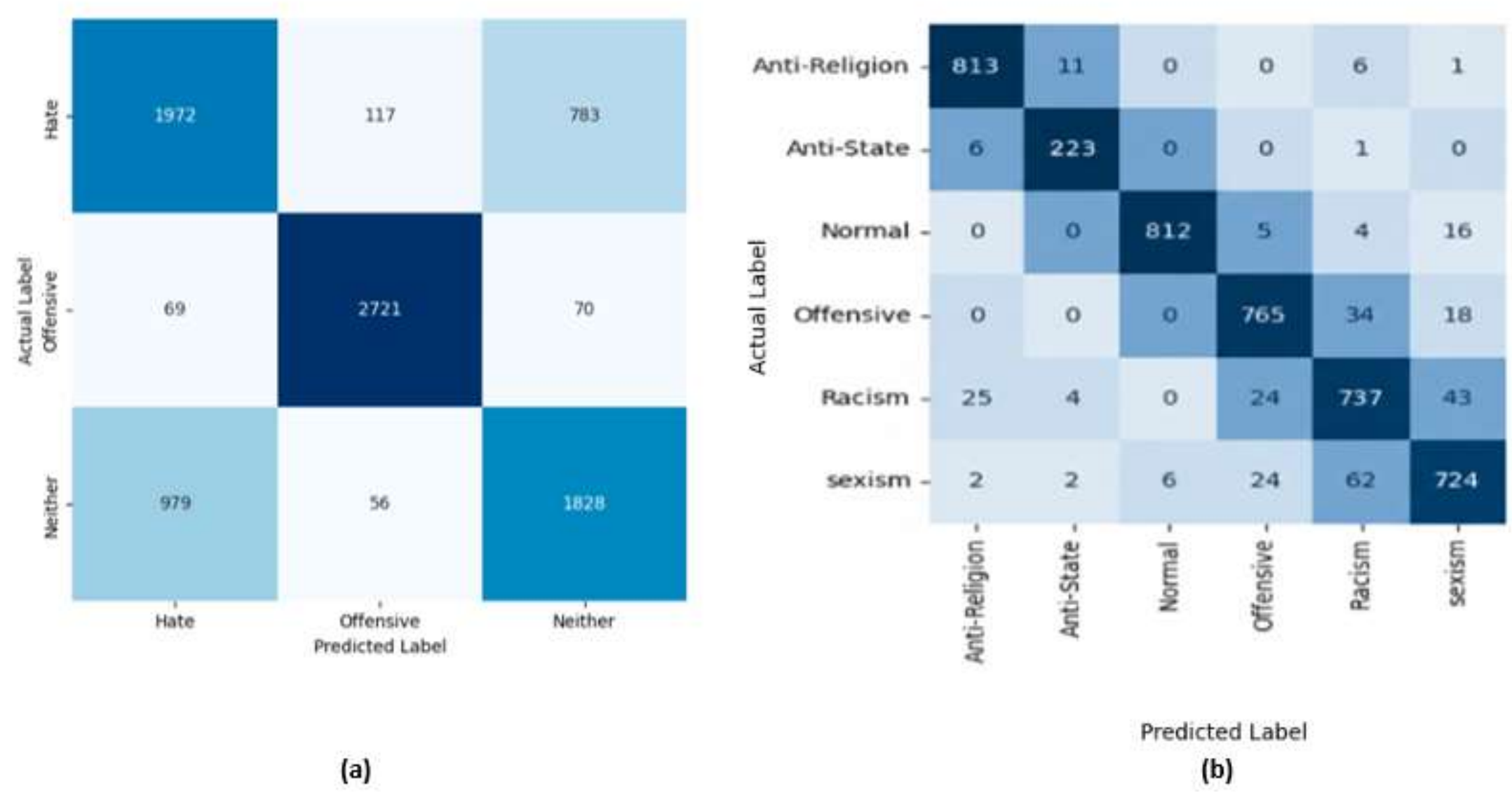

**Fig. 8**: Confusion matrix for (a) Davidson dataset, (b) SMHS dataset

**Table 6:** Comparison of classification performance for multi classification

| Dataset | Models | Accuracy (%) | Precision (%) | Recall (%) | F1-Score (%) |
|---|---|---|---|---|---|
| **Davidson dataset** | **[24]** | 92.42 | 91.00 | 92.00 | 91.00 |
| | **[31]** | 88.49 | 92.49 | 88.49 | 90.05 |
| | **[32]** | 93.10 | 93.50 | 93.70 | 93.60 |
| | ***Proposed*** | ***96.09*** | ***96.78*** | ***96.07*** | ***97.00*** |
| **SMHS dataset** | **[33]** | 90.00 | 91.00 | 90.00 | 90.00 |
| | ***Proposed*** | ***94.99*** | ***94.98*** | ***94.76*** | ***94.99*** |

For the SMHS dataset, the baseline study utilized DistilBERT embeddings and achieved 90% accuracy and F1-score, with precision and recall around 91%. While computationally efficient, it lacks sequential and attention-based processing, limiting its ability to discriminate between subtle class distinctions. The proposed model addresses this by combining contextual embeddings with BiLSTM and attention, achieving 94.99% F1-score and balanced precision (94.98%) and recall (94.76%), demonstrating more robust and consistent classification across all classes.

This completes the performance analysis of proposed framework and its comparison with standard benchmark studies for binary and multi-class classification tasks.

### 6.4 Explanations using LIME Model

This section discussed the interpretation of the predictive outcomes for the binary and multiclass classification settings using the Davidson and SMHS datasets. To better understand how the model makes its predictions, we applied LIME as an XAI technique. The process is completed by providing the LIME with multiple tweets for binary and multi-class classification taken from each dataset.

Considering Davidson dataset, the interpretation of the binary outcome is discussed here. LIME model assigned the label of Hate to a post, by giving it a score of 64%, whereas not-hate class label has a score of 36% as shown in **Fig.**

9. LIME highlights the word "*bitch*" as the most influential term contributing to this prediction. This word carries strong abusive meaning and has a high association with hateful content, which is why it strongly pushes the model toward the Hate class. In contrast, other words such as "*she*", "*is*", "*trying*", and "*to be*" contribute very little to the prediction because they are neutral and do not carry harmful meaning. This shows that the model's decision is mainly driven by semantically strong and offensive words rather than common grammatical words.

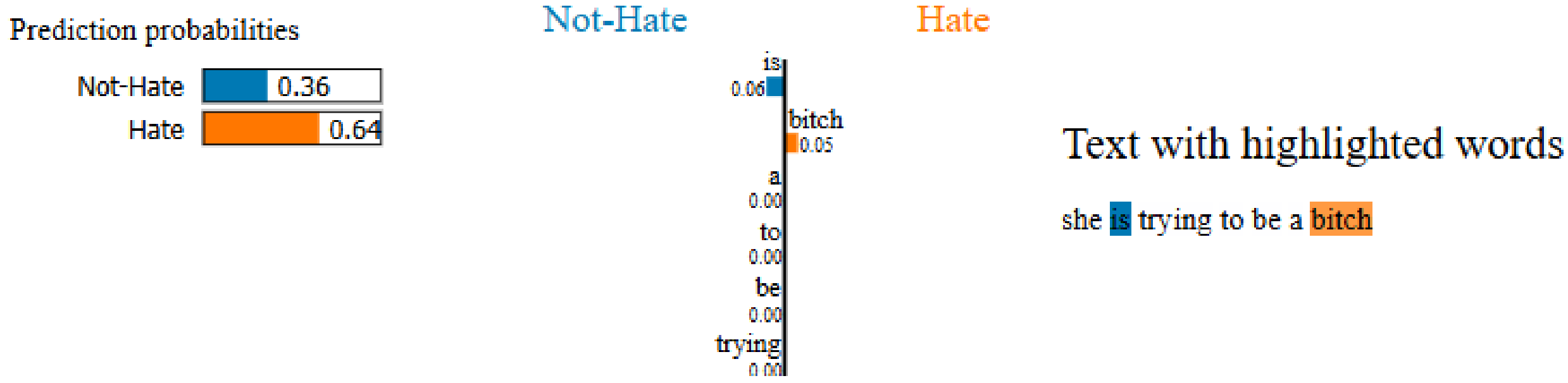

**Fig. 9**: LIME Visualization for Davidson dataset (binary classification)

For the interpretation of multi classes in Davidson dataset, a post is chosen and the response of LIME is presented in **Fig. 10**. The LIME classified the post into "Offensive Language" as it gave a score of 46% to this class which is the highest among others (27% and 26% for the class HS and Neither respectively). The word "*hoe*" is highlighted as the most important contributor to this class. This term is commonly used as an insult and is more strongly linked with "Offensive Language" than with explicit HS, which explains why the model favors the Offensive category. Other words such as "*John*", "*who*", and "*lookin*" have little influence on the prediction, as they are context-related but not semantically harmful.

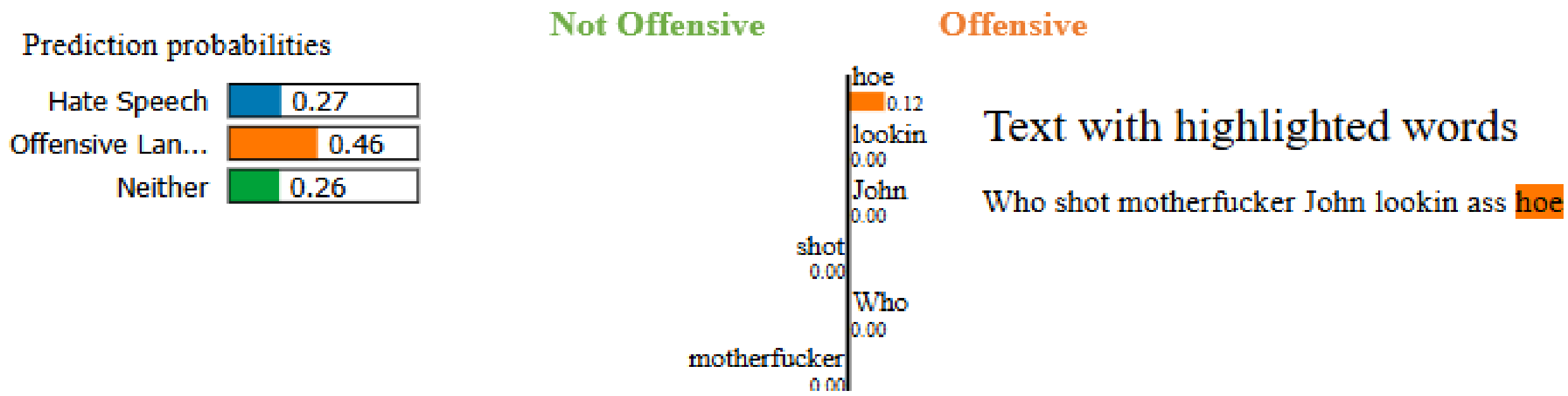

**Fig. 10**: LIME Visualization for Davidson dataset (multiclass classification)

The interpretation of outcome for the SMHS dataset is described here: Only one post is chosen here for binary and multi-class classification purposes. LIME predicts the given post as Hate class as shown in **Fig. 11**, which represents the binary classification setting. The Hate label is predicted with a probability of 63% and LIME highlights the words "*yall*", and "*hoes*" as the most influential contributors to this prediction. These words carry strong offensive or derogatory meaning and frequently appear in hate-related contexts within the dataset. As a result, they push the model toward the Hate class. Other words such as "*cuffing*", "*ain't*", "*never*", and "*have*" have minimal impact because they mainly serve as a structural or contextual role and do not strongly indicate hateful intent on their own.

For the multiclass classification task on the SMHS dataset, LIME describes why the model predicts the tweet as Offensive rather than categories such as Racism, Sexism, or Normal as show in ***Fig. 12***. The highlighted words "*yall*", "*niggas*", and "*cause*" again contributed the most to the prediction. Although these terms are offensive, they are not directed toward a specific protected group in a clearly targeted manner. This explains why the model assigns a higher probability to the Offensive class instead of Racism or Sexism. Words related to sentence flow, such as "*have*", "*never*", and "*ain't*", show very low contribution and therefore do not influence the final decision significantly.

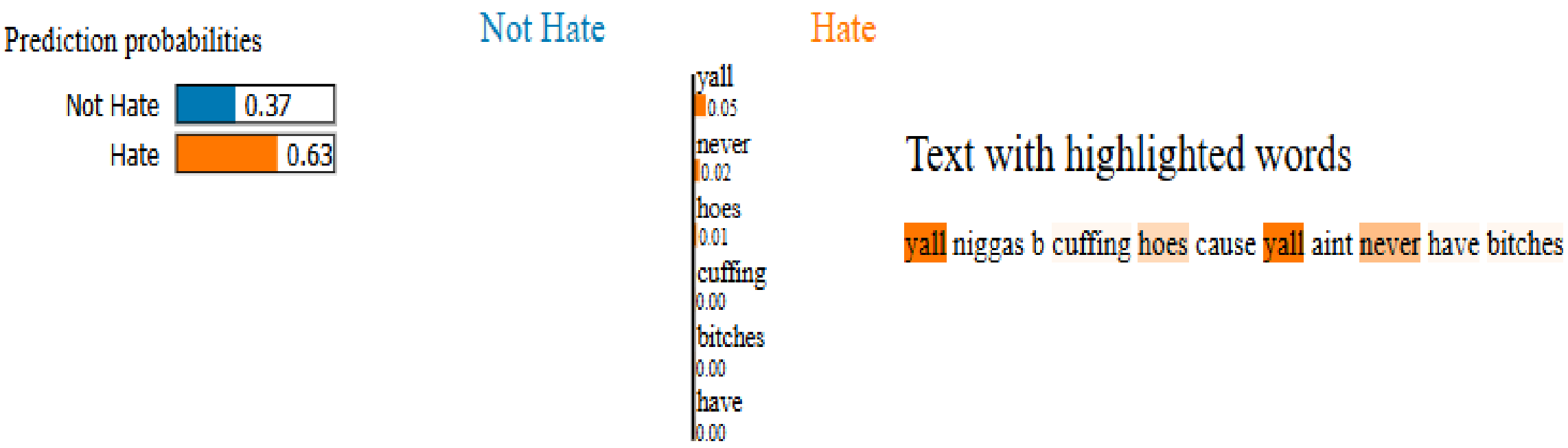

**Fig. 11**: LIME Visualization for SMHS dataset (binary classification)

Overall, the LIME interpretations across both datasets confirm that the models' predictive power is rooted in identifying high-impact, derogatory keywords rather than neutral grammatical structures. By consistently highlighting specific offensive terms as the primary drivers for both binary and multiclass labels, the XAI analysis validates that the models effectively distinguish harmful intent from general context. This transparency ensures that the classification process is both semantically grounded and aligned with human linguistic intuition.

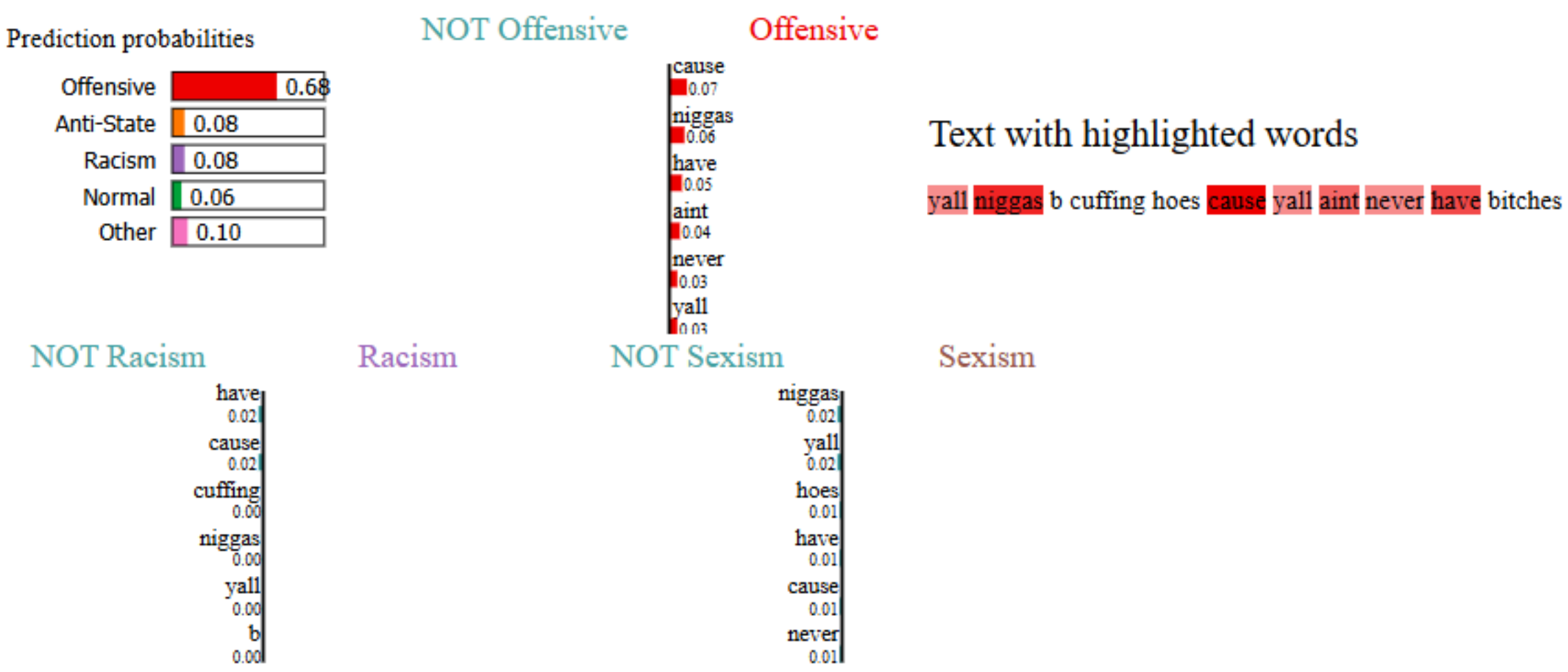

**Fig. 12**: LIME Visualization for SMHS dataset (multiclass classification)

**7. Conclusion and Future Directions**

A significant rise in HS and OL content make automated detection an important and challenging research problem. This challenge becomes more complex in multiclass and informal SM text, where linguistic variation and ambiguities are common. In this work, we proposed an explainable, multilevel DL based framework for HSD that leverages contextual transformer-based embeddings combined with sequential modeling. The datasets were passed through extensive pre-processing steps to filter out unnecessary information and remove noise. The model captures both global contextual information and local word-level dependencies, enabling effective discrimination between hateful, offensive, and neutral content. To improve transparency, LIME was employed to provide instance-level explanations, highlighting the most influential words contributing to each prediction. Experimental evaluation on the Davidson and SMHS datasets demonstrates benchmark performance in both binary and multiclass classification settings. For binary classification, the proposed framework achieved F1-scores of 96.78% and 99.53% on the Davidson and SMHS datasets. In the multi-class setting, it demonstrates F1-scores of 97.00% and 94.99% on the Davidson and SMHS datasets, respectively, outperforming existing baseline approaches both classification tasks (binary and multi-class). LIME-based interpretations further confirm that the predictions are driven by relevant linguistic cues rather than neutral or structural tokens, supporting the reliability and interpretability of the proposed approach. This indicates that the model consistently identifies semantically meaningful and abusive terms as key decision factors.

For future work, the framework can be extended to handle code-mixed and multilingual text, images and audio, which is prevalent in real-world SM data. Additionally, incorporating more advanced explainability techniques and performing comparative analysis across multiple interpretation methods may provide deeper insights into model behavior. Future research may also focus on fine-grained target identification and contextual intent modeling to improve practical deployment in content moderation systems.

**Statements and Declarations**

**Authorship Contributions**

Rameesha Zia: Visualization, Validation, Software, Investigation, Formal analysis, Data curation, Writing – original draft, Muhammad Shahid Iqbal Malik: Conceptualization, Methodology, Supervision, Formal analysis, Validation, Data curation, Writing – original draft, Writing – review & editing,

**Data Availability Statement**

The datasets used and/or analyzed during the current study are available from the corresponding author on reasonable request.

**Funding**

No funds, grants, or other support was received.

**Competing Interest Statement**

The authors have no relevant financial or non-financial interests to disclose.